\documentclass[a4paper,fleqn]{cas-dc}

\usepackage[numbers]{natbib}
\usepackage{microtype}

\usepackage{hyperref}
\hypersetup{colorlinks=true, citecolor=blue, linkcolor=blue, urlcolor=blue}
\usepackage[commandnameprefix=always]{changes}
\usepackage{todonotes}
\usepackage{algorithm}
\usepackage{algorithmic}
\usepackage{xcolor}
\usepackage{float} 
\usepackage{tabularx}
\usepackage{caption}

\usepackage{comment}
\def\tsc#1{\csdef{#1}{\textsc{\lowercase{#1}}\xspace}}
\tsc{WGM}
\tsc{QE}

\newcommand{\figref}[1]{\hyperref[#1]{Figure\ref*{#1}}}

\newcommand{\mytabref}[1]{\hyperref[#1]{Table\ref*{#1}}}

\newcommand{\myalgref}[1]{\hyperref[#1]{Algorithm\ref*{#1}}}

\newcommand{\myeqref}[1]{\hyperref[#1]{Equation\ref*{#1}}}

\begin{document}
\let\WriteBookmarks\relax
\def\floatpagepagefraction{1}
\def\textpagefraction{.001}
\let\printorcid\relax
\shorttitle{Armin Abdollahi et al. Advanced Predictive Modeling for Enhanced Mortality Prediction in ICU Stroke Patients Using Clinical Data}   

\shortauthors{Armin Abdollahi et al.}

\title[mode = title]{Advanced Predictive Modeling for Enhanced Mortality Prediction in ICU Stroke Patients Using Clinical Data}  

\author[a]{Armin Abdollahi}
\author[b]{Negin Ashrafi}
\author[b]{Maryam Pishgar}
\cormark[1]

\address[a]{Department of Electrical and Computer Engineering, University of Southern California (USC), 3740  McClintock Ave, Los Angeles, CA 90089, USA}

\address[b]{Department of Industrial and Systems Engineering, University of Southern California (USC), 3715  McClintock Ave, Los Angeles, CA 90089, USA}

\cortext[cor1]{Corresponding author.\\\hspace*{2em} \textit{E-mail address:} \href{mailto:pishgar@usc.edu}{pishgar@usc.edu} (Maryam Pishgar).}
\begin{abstract}
\noindent\textit{Background:} Stroke is second-leading cause of disability and death among adults. Approximately 17 million people suffer from a stroke annually, with about 85\% being ischemic strokes. Predicting mortality of ischemic stroke patients in intensive care unit (ICU) is crucial for optimizing treatment strategies, allocating resources, and improving survival rates.

\noindent\textit{Methods:} We acquired data on ICU ischemic stroke patients from MIMIC-IV database, including diagnoses, vital signs, laboratory tests, medications, procedures, treatments, and clinical notes. Stroke patients were randomly divided into training (70\%, n=2441), test (15\%, n=523), and validation (15\%, n=523) sets. To address data imbalances, we applied Synthetic Minority Over-sampling Technique (SMOTE). We selected 30 features for model development, significantly reducing feature number from 1095 used in the best study. We developed a deep learning model to assess mortality risk and implemented several baseline machine learning models for comparison.

\noindent\textit{Results:} XGB-DL model, combining XGBoost for feature selection and deep learning, effectively minimized false positives. Model’s AUROC improved from 0.865 (95\% CI: 0.821 - 0.905) on first day to 0.903 (95\% CI: 0.868 - 0.936) by fourth day using data from 3,646 ICU mortality patients in the MIMIC-IV database with 0.945 AUROC (95\% CI: 0.944 - 0.947) during training. Although other ML models also performed well in terms of AUROC, we chose Deep Learning for its higher specificity. 

\noindent\textit{Conclusions:} Through enhanced feature selection and data cleaning, proposed model demonstrates a 13\% AUROC improvement compared to existing models while reducing feature number from 1095 in previous studies to 30.
\end{abstract}


\maketitle
\section{Background}

The intensive care unit (ICU) is a structured system designed to care for critically ill patients, offering intensive and specialized medical and nursing services, advanced monitoring capabilities, and multiple physiological organ support modalities to sustain life during periods of severe organ system failure \cite{ref1}. In the United States, stroke is a leading cause of death and disability, underscoring the critical importance of ICU care for stroke patients \cite{ref2}.

Ischemic stroke occurs when blood flow to the brain is blocked or reduced, posing significant health risks \cite{ref3}. In recent years, approximately 13.7 million people suffer strokes annually, with 5.8 million resulting in death, of which 70\% are ischemic strokes \cite{ref4}. The large number of stroke patients significantly exacerbates the challenge of proper ICU resource allocation, particularly during the COVID-19 era. Logistically, there is a severe shortage of equipment and medications (such as ventilators and syringe pumps), while the number of patients far exceeds hospital capacity, preventing medical staff from providing timely treatment \cite{ref5}. Stroke patients requiring intensive care are at extremely high risk of short-term death, although this risk diminishes with increased survival time following ICU admission \cite{ref6}.

ICUs also cater to patients with other critical conditions. For instance, machine learning models have been developed to predict in-hospital mortality for ICU patients with heart failure, demonstrating the utility of advanced algorithms in critical care settings \cite{ref38}. Similarly, deep learning models have been utilized to predict mortality in mechanically ventilated ICU patients, highlighting the significance of predictive analytics in managing complex ICU cases \cite{ref24}.

From a genetic standpoint, hereditary conditions such as hypertension and diabetes may be passed down through familial bloodlines, increasing the potential risk of stroke in otherwise healthy individuals \cite{ref7, ref8, ref9}. Alternatively, harmful lifestyle practices, such as smoking and lack of exercise, are also significant factors leading to the frequent occurrence of strokes \cite{ref10}.

With the advent of machine learning, algorithms have been increasingly applied to various disease prediction models \cite{ref11, ref45}. Compared to traditional statistical methods, machine learning can rapidly process numerous features, consider more permutations, and enhance prediction accuracy \cite{ref12, ref13}. A substantial proportion of machine learning models developed for disease analysis focus on stroke patients \cite{ref14, ref15}. These mortality prediction models for stroke patients are widely used in clinical medicine to provide timely warnings to ICU doctors and to facilitate the efficient allocation of medical resources \cite{ref16, ref17}.

Neural network models and deep learning represent the forefront of artificial intelligence, transforming how machines process information and make decisions \cite{ref36, ref18, ref37}. Neural networks mimic the interconnected neurons in the brain to process complex data, and one of their key strengths is the ability to learn intricate patterns and relationships from data without explicit programming \cite{ref19, ref20}. 

The primary objective of this research was to develop a deep learning model for predicting the mortality of ischemic stroke patients using ICU patient data from the MIMIC-IV database. Compared to the primary reference article, we employed feature selection to reduce the number of predictor variables while improving the accuracy of the results. The predictive model was developed following the guidelines of the Transparent Reporting of a Multivariable Prediction Model for Individual Prognosis or Diagnosis (TRIPOD) initiative \cite{ref47}.

\section{Methodologies}
\subsection{Data Source and Study Design}

Our study utilized the Medical Information Mart for Intensive Care (MIMIC-IV) database, a contemporary electronic health record dataset resulting from a collaboration between Beth Israel Deaconess Medical Center (BIDMC) and the Massachusetts Institute of Technology (MIT) \cite{ref21}. Specific data, including patient diagnoses, vital signs, laboratory tests, medications, procedures, treatments, and de-identified free-text clinical notes, were extracted from the MIMIC-IV database to cover specific patient cohorts. The MIMIC-IV database was chosen because it provides an extensive amount of real ICU patient data and, compared to MIMIC-III, offers more accurate updates and organizes the data into a modular structure. This facilitates the formulation of hypotheses for more comprehensive research problems and their application in clinical medicine. After data extraction, preprocessing is essential to ensure high data quality and to organize it into a format suitable for analysis by machine learning algorithms. The data within the MIMIC-IV database serves as a robust foundation for research endeavors, effectively supporting the development of deep learning models and benefiting clinical medical personnel.

\subsection{\textit{Patient extraction}}

\begin{figure*}[h]
    \centering
    \includegraphics[width=0.8\linewidth]{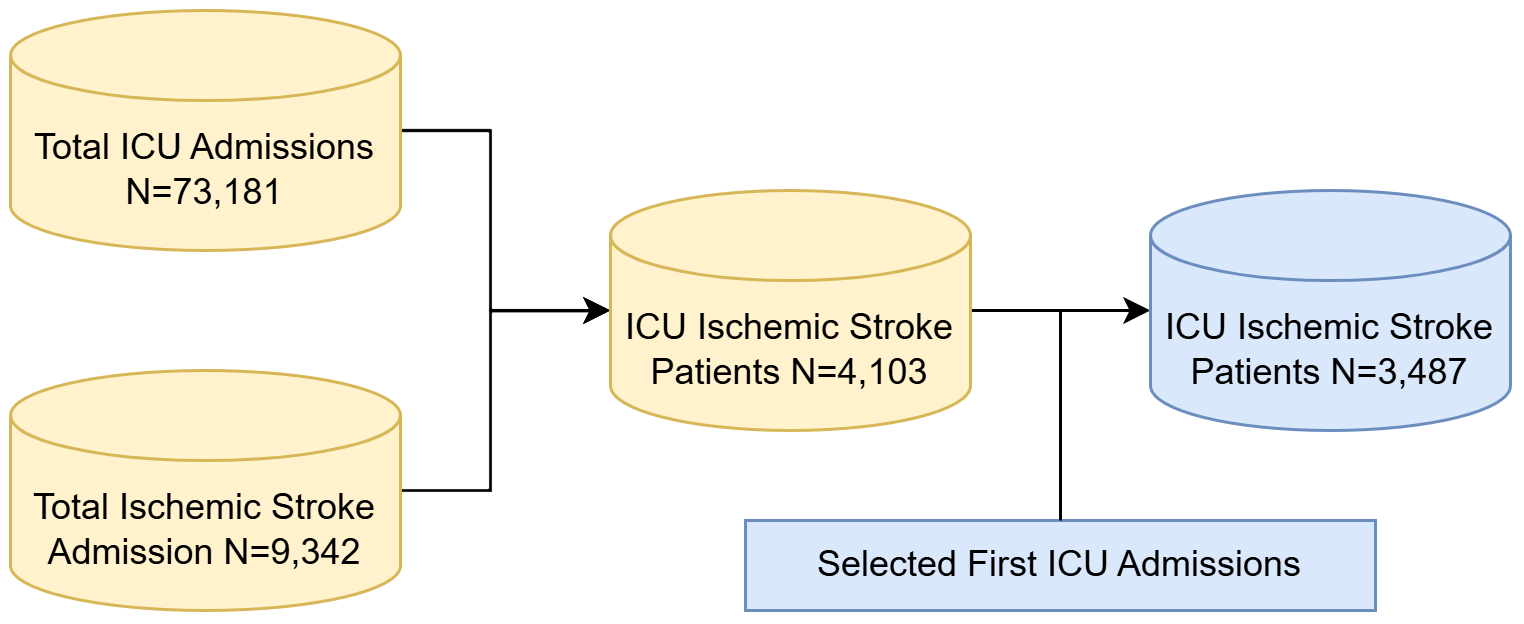}
    \captionsetup{justification=centering}
    \caption{Flow diagram of the selection process of patients.}
    \label{fig:1}
\end{figure*}

Our research focused on predicting mortality in ICU patients with ischemic stroke. \figref{fig:1} illustrates the patient extraction process. Initially, we selected 73,181 ICU patients and 9,342 ischemic stroke patients from the database. From these, we identified 4,103 patients with ischemic stroke in the ICU. Ultimately, we included the first ICU admission for each patient, resulting in a total of 3,487 patients who met the established inclusion criteria for the final analysis.

We exclusively included first-time ICU admissions for each patient in this study to maintain the consistency and reliability of the dataset. By focusing on first-time admissions, we aimed to eliminate potential confounding factors associated with multiple admissions, such as varying treatment responses, changes in health conditions, or differences in care practices across different ICU stays. This approach helps to ensure that the predictions made by our model are based on the initial severity and characteristics of the patients' conditions, rather than being influenced by previous ICU experiences or interventions.

\subsection{\textit{Data processing}}

For the dataset used in the research, a total of 1,295 features were initially considered. The input features are shown with $\mathbf{X}$ and $\mathbf{X}_{\text{initial}} \in \mathbb{R}^{n \times 1295}$ shows the dimension of the input where n is the number of rows in the dataset. We then followed the \myeqref{eq:filter_nan} to eliminate features containing more than 50\% NaN values and using expert opinion to reduce the number of features to 144 that might be related to the target variable. For the retained features, we imputed missing values with the median value. Additionally, we used \myeqref{eq:min_max_scaling} to normalize the numerical values to standardize the scales and improve convergence.

\begin{equation}
\mathbf{X}_{\text{filtered}} = \{ \mathbf{x}_i \in \mathbf{X}_{\text{initial}} \mid \text{NaN}(\mathbf{x}_i) \leq 0.5 \times n \}
\label{eq:filter_nan}
\end{equation}

\begin{equation}
\mathbf{X}_{\text{scaled}} = 2 \times \frac{\mathbf{X}_{\text{filtered}} - \min(\mathbf{X}_{\text{filtered}})}{\max(\mathbf{X}_{\text{filtered}}) - \min(\mathbf{X}_{\text{filtered}})} - 1
\label{eq:min_max_scaling}
\end{equation}

\subsection{\textit{Feature selection}}

We used XGBoost, and LASSO, along with expert opinion, to select 30 key predictors for subsequent analysis. Subject IDs, lab event IDs, and ICU stay IDs serve as unique identifiers for patients, laboratory events, and ICU admissions, respectively. All physiological test indicators and disease diagnoses were referenced using ICD-9 codes. \mytabref{tab:1} presents the proposed 30 features, including:

(I) GCS - Eye Opening: The patient's level of consciousness based on their response to stimuli.
(II) O2 Flow (L/min): The rate at which oxygen is administered to the patient.
(III) GCS - Verbal Response: The patient's level of consciousness based on their verbal response to stimuli.
(IV) GCS - Motor Response: The patient's level of consciousness based on their motor response to stimuli.
(V) Intravenous / IV Access Prior to Admission: Indicates whether the patient had intravenous access established before ICU admission.
(VI) Ventilator Type: Specifies the type of ventilator used for respiratory support.
(VII) Anion Gap: The difference between measured cations and anions in the blood.
(VIII) Insulin Pump: Indicates whether the patient is using an insulin pump for administering insulin.
(IX) Arterial CO2 Pressure (mmHg): The partial pressure of carbon dioxide in arterial blood.
(X) Respiratory Rate (Total) (insp/min): The total respiratory rate, measured in breaths per minute.
(XI) Braden Nutrition: Assessing a patient's risk for pressure ulcers related to nutrition.
(XII) O2 Saturation Pulseoxymetry Alarm - High (\%): The high alarm threshold for oxygen saturation as measured by pulse oximetry.
(XIII) ST Segment Monitoring On: Indicates whether ST segment monitoring is activated.
(XIV) Braden Mobility: Assessing a patient's risk for pressure ulcers related to mobility.
(XV) Marital Status: The patient's marital status.
(XVI) HCO3 (serum): The concentration of bicarbonate ions in the blood serum.
(XVII) Chloride (serum): The concentration of chloride ions in the blood serum.
(XVIII) TCO2 (calc) Arterial: The calculated total carbon dioxide content in arterial blood.
(XIX) Creatinine: The concentration of creatinine in the blood.
(XX) O2 Saturation Pulseoxymetry (\%): Oxygen saturation as measured by pulse oximetry.
(XXI) Base Excess: The amount of excess or deficit of bases (bicarbonate) in the blood.
(XXII) BUN: Blood urea nitrogen.
(XXIII) Self ADL: Self-assessed activities of daily living.
(XXIV) RDW: Red blood cell distribution width.
(XXV) Respiratory Rate (spontaneous) (insp/min): The respiratory rate during spontaneous breathing.
(XXVI) Red Blood Cells: The concentration of red blood cells in the blood.
(XXVII) INR (PT): International normalized ratio.
(XXVIII) Braden Friction/Shear: Assessing a patient's risk for pressure ulcers related to friction and shear.
(XXIX) Daily Weight (kg): The patient's weight measured daily.
(XXX) Alarms On: Indicates whether alarms are activated.

\begin{table*}[t]
\renewcommand{\arraystretch}{1} 
\caption{Feature List after applying the XGBoost feature important techniques, expert opinion, and literature review}
\label{tab:1}
\begin{tabularx}{\textwidth}{|c|X|c|X|}
\hline
\textbf{No} & \textbf{Feature} & \textbf{No} & \textbf{Feature} \\
\hline
1 & GCS - Eye Opening & 16 & HCO3 (serum) \\
\hline
2 & O2 Flow (L/min) & 17 & Chloride (serum) \\
\hline
3 & GCS - Verbal Response & 18 & TCO2 (calc) Arterial \\
\hline
4 & GCS - Motor Response & 19 & Creatinine \\
\hline
5 & Intravenous / IV access prior to admission & 20 & O2 saturation pulseoxymetry (\%) \\
\hline
6 & Ventilator Type & 21 & Base Excess \\
\hline
7 & Anion Gap & 22 & BUN \\
\hline
8 & Insulin pump & 23 & Self ADL \\
\hline
9 & Arterial CO2 Pressure (mmHg) & 24 & RDW \\
\hline
10 & Respiratory Rate (Total) (insp/min) & 25 & Respiratory Rate (spontaneous) (insp/min) \\
\hline
11 & Braden Nutrition & 26 & Red Blood Cells \\
\hline
12 & O2 Saturation Pulseoxymetry Alarm - High & 27 & INR(PT) \\
\hline
13 & ST Segment Monitoring On & 28 & Braden Friction/Shear \\
\hline
14 & Braden Mobility & 29 & Daily Weight (kg) \\
\hline
15 & marital\_status & 30 & Alarms On \\
\hline
\end{tabularx}
\end{table*}

In our research, we applied two models, XGBoost and LASSO, for feature selection, each offering unique benefits. XGBoost is a scalable tree boosting system that excels in achieving high predictive accuracy across various domains, making it a popular choice in machine learning applications \cite{ref22}. It also includes regularization parameters that help prevent overfitting while capturing complex relationships in the data. Furthermore, XGBoost's advanced feature selection capabilities enable the identification of the most relevant predictors while minimizing noise, thereby enhancing model interpretability and generalization performance. Despite its widespread adoption, XGBoost's complex ensemble of decision trees can pose challenges in model interpretation and fine-tuning. The parameters of the used XGBoost are summarized in the \mytabref{tab:xgb_params}.

\begin{table}[h]
    \centering
    \renewcommand{\arraystretch}{1} 
    \setlength{\tabcolsep}{10pt} 
    \caption{XGBoost model parameters used in this paper and their values}
    \label{tab:xgb_params}
    \begin{tabular}{|l|l|}
        \hline
        \textbf{Parameter} & \textbf{Value} \\ \hline
        n\_estimators      & 150             \\ \hline
        base\_score        & 0.5             \\ \hline
        learning\_rate     & 0.1             \\ \hline
        max\_depth         & 5               \\ \hline
        min\_child\_weight & 1               \\ \hline
        gamma              & 0               \\ \hline
        subsample          & 1               \\ \hline
        colsample\_bytree  & 1               \\ \hline
        colsample\_bylevel & 1               \\ \hline
        colsample\_bynode  & 1               \\ \hline
        reg\_alpha         & 0               \\ \hline
        reg\_lambda        & 1               \\ \hline
        scale\_pos\_weight & 1               \\ \hline
        max\_delta\_step   & 0               \\ \hline
    \end{tabular}
\end{table}

LASSO, a widely used regression technique, is renowned for its ability to perform feature selection and enhance model interpretability \cite{ref27}. By shrinking regression coefficients towards zero, LASSO encourages sparsity in the model, effectively identifying the most influential predictors \cite{ref28}. However, LASSO's variable selection may be biased towards those with higher coefficients, potentially overlooking important but smaller effects \cite{ref29}. Considering the strengths and weaknesses of each model, we integrated the features identified by both models into the training of our predictive model. The LASSO parameters are shown in \mytabref{tab:lasso_params}. We determined the ultimate feature selection model based on the accuracy, precision, sensitivity, F1-score, and specificity of the parameters obtained. 

\begin{table}[h]
    \centering
    \renewcommand{\arraystretch}{1} 
    \setlength{\tabcolsep}{10pt} 
    \caption{LASSO model parameters used in this paper and their values}
    \label{tab:lasso_params}
    \begin{tabular}{|l|l|}
        \hline
        \textbf{Parameter} & \textbf{Value} \\ \hline
        alpha              & 0.005 \\ \hline
        max\_iter          & 900            \\ \hline
        tol                & 0.0001          \\ \hline
        selection          & cyclic        \\ \hline
    \end{tabular}
\end{table}

\subsection{\textit{Ablation process}}

To assess whether the 30 selected features would adversely affect model performance, we gradually eliminated variables that negatively impacted the model. We evaluated performance on the validation set by calculating the 95\% CI of the AUROC. We sequentially deleted one variable at a time, repeating the process until further deletions no longer improved performance. This method filters out non-contributing variables, enhancing model accuracy. After this process, we found all 30 features positively influenced performance, so we decided to retain all features. This algorithm is summarized in \myalgref{alg:feature_selection}.

\begin{algorithm}
\caption{Feature Selection Using AUROC Evaluation}
\label{alg:feature_selection}
\begin{algorithmic}[1]
\REQUIRE Initial set of features $\mathbf{X}_{\text{initial}} \in \mathbb{R}^{n \times 30}$, Validation dataset $\mathbf{X}_{\text{val}}$, Target variable $\mathbf{y}_{\text{val}}$
\STATE $\mathbf{X}_{\text{current}} \gets \mathbf{X}_{\text{initial}}$
\STATE $\text{AUROC}_{\text{best}} \gets \text{CalculateAUROC}(\mathbf{X}_{\text{current}}, \mathbf{y}_{\text{val}})$
\STATE $\text{Improvement} \gets \text{True}$
\WHILE{$\text{Improvement}$}
    \STATE $\text{Improvement} \gets \text{False}$
    \FOR{each feature $x_i$ in $\mathbf{X}_{\text{current}}$}
        \STATE $\mathbf{X}_{\text{temp}} \gets \mathbf{X}_{\text{current}} \setminus \{x_i\}$
        \STATE $\text{AUROC}_{\text{temp}} \gets \text{CalculateAUROC}(\mathbf{X}_{\text{temp}}, \mathbf{y}_{\text{val}})$
        \IF{$\text{AUROC}_{\text{temp}} > \text{AUROC}_{\text{best}}$}
            \STATE $\text{AUROC}_{\text{best}} \gets \text{AUROC}_{\text{temp}}$
            \STATE $\mathbf{X}_{\text{current}} \gets \mathbf{X}_{\text{temp}}$
            \STATE $\text{Improvement} \gets \text{True}$
        \ENDIF
    \ENDFOR
\ENDWHILE
\STATE $\mathbf{X}_{\text{final}} \gets \mathbf{X}_{\text{current}}$
\RETURN $\mathbf{X}_{\text{final}}$
\end{algorithmic}
\end{algorithm}

\subsection{\textit{Modeling}}

The dataset was imbalanced, with a survival-to-death ratio of 4:1 (1935:505). To address this issue, we implemented the Synthetic Minority Over-Sampling Technique (SMOTE)\cite{ref46}. Additionally, the train\_test\_split method was used to divide the dataset into training, testing, and validation sets(70/15/15). We developed a novel deep learning neural network to predict mortality in ICU patients with ischemic stroke. For comparison, we established four baseline machine learning models: Random Forest, Logistic Regression, XGBoost, and LightGBM \cite{ref30, ref31, ref22, ref33}. To ensure the robustness and reliability of our predictive models, we implemented five-fold cross-validation to minimize the impact of a single dataset split and provide a comprehensive evaluation of the models' generalizability and stability."

The choice of a deep learning model over traditional machine learning models was motivated by the need to handle the complex, high-dimensional nature of ICU patient data. Deep learning models are particularly well-suited for capturing non-linear relationships and interactions among multiple features, which are common in healthcare data. Compared to other models, such as Random Forest, Logistic Regression, XGBoost, and LightGBM, deep learning can better learn from the rich, high-dimensional data. Additionally, our deep learning model showed superior performance in preliminary tests, achieving higher AUROC and specificity. This indicates a better ability to reduce false positives and accurately predict patient outcomes, which is crucial in critical care settings. Therefore, we selected deep learning as the primary model for its potential to provide more precise and reliable mortality predictions in ICU stroke patients.

\begin{figure}[!htb]
    \centering
    \includegraphics[width=1\linewidth]{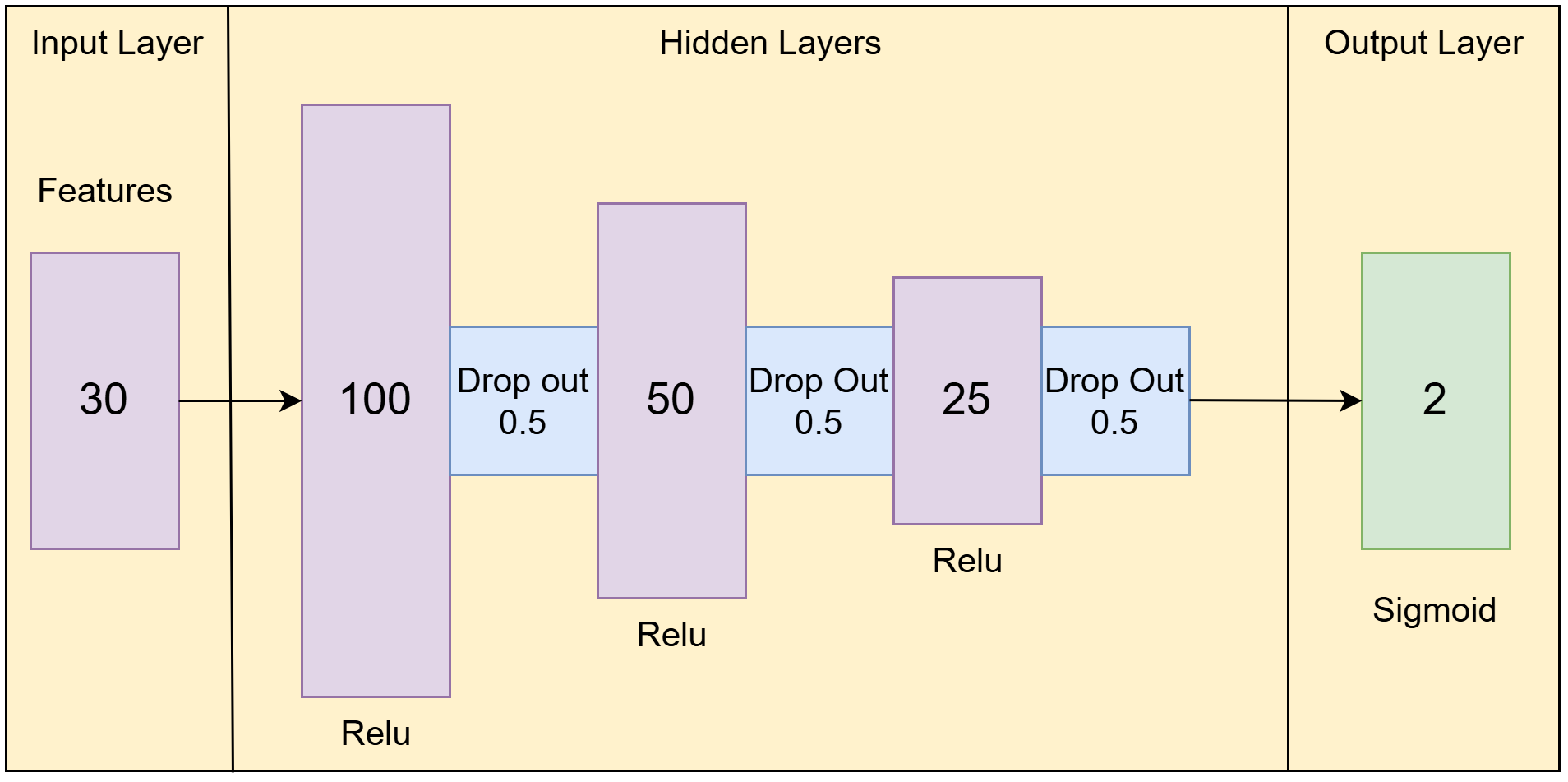}
    \captionsetup{justification=centering}
    \caption{This figure illustrates the neural network structure used, consisting of three hidden layers with 100, 50, and 25 neurons, respectively, each incorporating a dropout rate of 0.5. The model was trained using 30 input features.}
    \label{fig:2}
\end{figure}

\figref{fig:2} illustrates the architecture of our deep learning neural network (NN) model. This model consists of a fully connected NN with an initial layer of 30 dimensions, followed by a batch normalization (BN) layer for input normalization, enhancing the model's stability \cite{ref34}. The batch normalization process is defined in \myeqref{eq:batch}, here $\mu_b$ is the mean value of the batch and $\sigma_b$ is the standard deviation of batch, and the $\epsilon$ is a small constant to avoid division by zero. The model includes three hidden layers, each employing the rectified linear unit (ReLU) activation function which is defined by \myeqref{eq:relu}. Dropout layers were utilized between these hidden layers to mitigate overfitting. The number of neurons decreases from 100 in the first hidden layer to 25 in the third hidden layer. The output layer contains a single neuron, using the sigmoid activation function given in \myeqref{eq:sig} for binary classification, producing output probabilities ranging from 0 to 1. The model was trained using the SGD optimizer, with binary\_crossentropy as the loss function and AUROC as the evaluation metric. The training process ran for 100 epochs with a batch size of 32. This series of operations enhances the model's ability to distinguish between positive and negative cases.

\begin{equation}
\hat{x}_i = \frac{x_i - \mu_B}{\sqrt{\sigma_B^2 + \epsilon}}
\label{eq:batch}
\end{equation}

\begin{equation}
\text{ReLU}(x) = \max(0, x)
\label{eq:relu}
\end{equation}

\begin{equation}
\sigma(x) = \frac{1}{1 + e^{-x}}
\label{eq:sig}
\end{equation}

The best model was selected based on its performance on the validation set. Additionally, we calculated accuracy, precision, sensitivity, F1-score, and specificity to evaluate our models' performance.

\section{RESULTS}

\subsection{\textit{Cohort Comparison}}

We extracted data for 3,646 ICU patients from the MIMIC-IV database for the development of our model. The cohort was then randomly divided into three subsets: 2,440 patients were allocated to the training set, 682 patients to the test set, and 524 patients to the validation set. The training and validation sets were used to train the models, and the model that achieved the highest AUROC value was selected as the optimal predictive model for further evaluation on the test set. \mytabref{tab:cohort} provides a comprehensive comparison of demographic and clinical characteristics between the training cohort (N=2440) and the validation cohort (N=524). 

\begin{table*}[htp]
\centering
\renewcommand{\arraystretch}{0.9} 
\caption{Comparison of Train and Validation Cohorts. Values from row 9 to row 20 are presented as mean (standard deviation). Some patients' race information is unknown.}
\label{tab:cohort}
\begin{tabularx}{1\textwidth}{|c|X|c|c|c|}
\hline
\textbf{No} & \textbf{Feature} & \textbf{Train Cohort (N=2440)} & \textbf{Validation Cohort (N=524)} & \textbf{P} \\
\hline
1 & Gender [M F] & [1293 1147] & [273 251] & 0.788 \\
\hline
2 & Target [Survive Death] & [1935 505] & [421 103] & 1.00 \\
\hline
3 & Race - White & 1690 (54.12\%) & 369 (55.10\%) & 0.789 \\
\hline
4 & Race - African American & 193 (6.17\%) & 40 (5.98\%) & 0.789 \\
\hline
5 & Race - Hispanic/Latino & 92 (2.94\%) & 12 (1.79\%) & 0.789 \\
\hline
6 & Race - Asian & 64 (2.05\%) & 14 (2.09\%) & 0.789 \\
\hline
7 & Race - American Indian/Alaska Native & 7 (0.22\%) & 2 (0.30\%) & 0.789 \\
\hline
8 & Race - Other & 174 (5.57\%) & 36 (5.61\%) & 0.789 \\
\hline
9 & Age & 68.01 (15.30) & 68.93 (15.43) & 0.625 \\
\hline
10 & GCS - Eye Opening & 2.92 (1.13) & 2.91 (1.15) & 0.399 \\
\hline
11 & O2 Flow (L/min) & 5.59 (5.73) & 5.37 (4.47) & 0.117 \\
\hline
12 & GCS - Verbal Response & 3.05 (1.74) & 3.10 (1.75) & 0.055 \\
\hline
13 & GCS - Motor Response & 4.99 (1.48) & 4.96 (1.51) & 0.070 \\
\hline
14 & Intravenous & 0.55 (0.50) & 0.51 (0.50) & 0.648 \\
\hline
15 & Ventilator Type & 1.02 (0.23) & 1.00 (0.56) & 0.590 \\
\hline
16 & Anion Gap & 14.24 (3.41) & 14.24 (3.62) & 0.999 \\
\hline
17 & Insulin pump & 0.00 (0.06) & 0.00 (0.00) & 0.605 \\
\hline
18 & Arterial CO2 Pressure (mmHg) & 38.74 (7.42) & 39.53 (9.00) & 0.410 \\
\hline
19 & Respiratory Rate (Total) (insp/min) & 18.95 (13.24) & 18.38 (4.44) & 0.101 \\
\hline
20 & Braden Nutrition & 2.38 (0.60) & 2.38 (0.59) & 0.06 \\
\hline
\end{tabularx}
\end{table*}

Key demographics such as age, gender, and race exhibit similar distributions across both cohorts, suggesting consistency and the potential for generalizability of the findings. Clinical parameters, including various GCS scores, oxygen flow rates, and other medical metrics, are compared with their mean values and standard deviations for each group. All p-values are greater than 0.05, indicating no statistically significant differences in these parameters between the cohorts, reinforcing the validation cohort's reliability as a representative sample for further research analysis or model validation.

\subsection{\textit{Ablation Study on Variable}}

\begin{figure*}[!htb]
    \centering
    \includegraphics[width=0.9\linewidth]{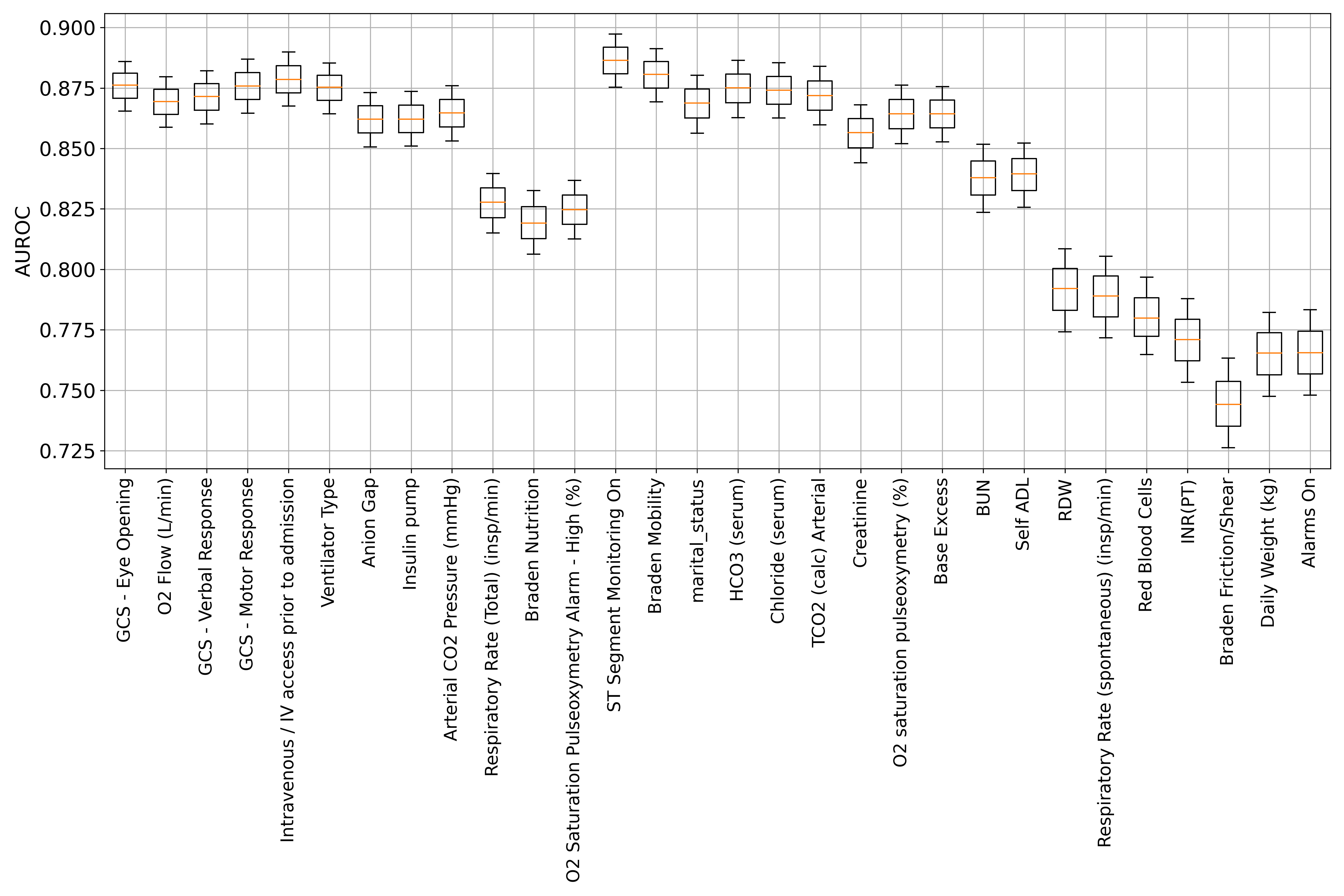}
    \caption{This figure presents the ablation study conducted for this paper. The upper and lower parts of each box plot represent the high and low ranges of the confidence interval, respectively, while the middle point indicates the AUROC.}
    \captionsetup{justification=centering}
    \label{fig:3}
\end{figure*}

The ablation study shown in \figref{fig:3} demonstrates that removing any feature negatively impacts the model's AUROC. None of the values in the figure surpass the original model, which includes all 30 features, and has an AUROC of 0.89.

This superior performance indicates that the baseline configuration already optimally captures the necessary predictive elements. The graphical results from the study show that subsequent ablations, which involve the systematic removal of features such as GCS Eye Opening, GCS Verbal Response, and other clinical variables, do not lead to an increase in AUROC values. In fact, in each instance where a feature is removed, the AUROC tends to decrease or remain unchanged compared to the baseline. This finding underscores that the current feature set within the baseline model is integral to its predictive success. Any removal of these features would not contribute positively to the model's performance; therefore, maintaining the existing feature composition is advisable. These results validate the robustness of the baseline model and suggest that the included features collectively enhance the model’s ability to accurately predict outcomes, negating the necessity for further adjustments or simplifications in the feature set. This stability in model performance with the existing features supports their continued use without modification for optimal results.

\subsection{\textit{Evaluation results}}

\begin{figure*}[h]
    \centering
    \includegraphics[width=1\linewidth]{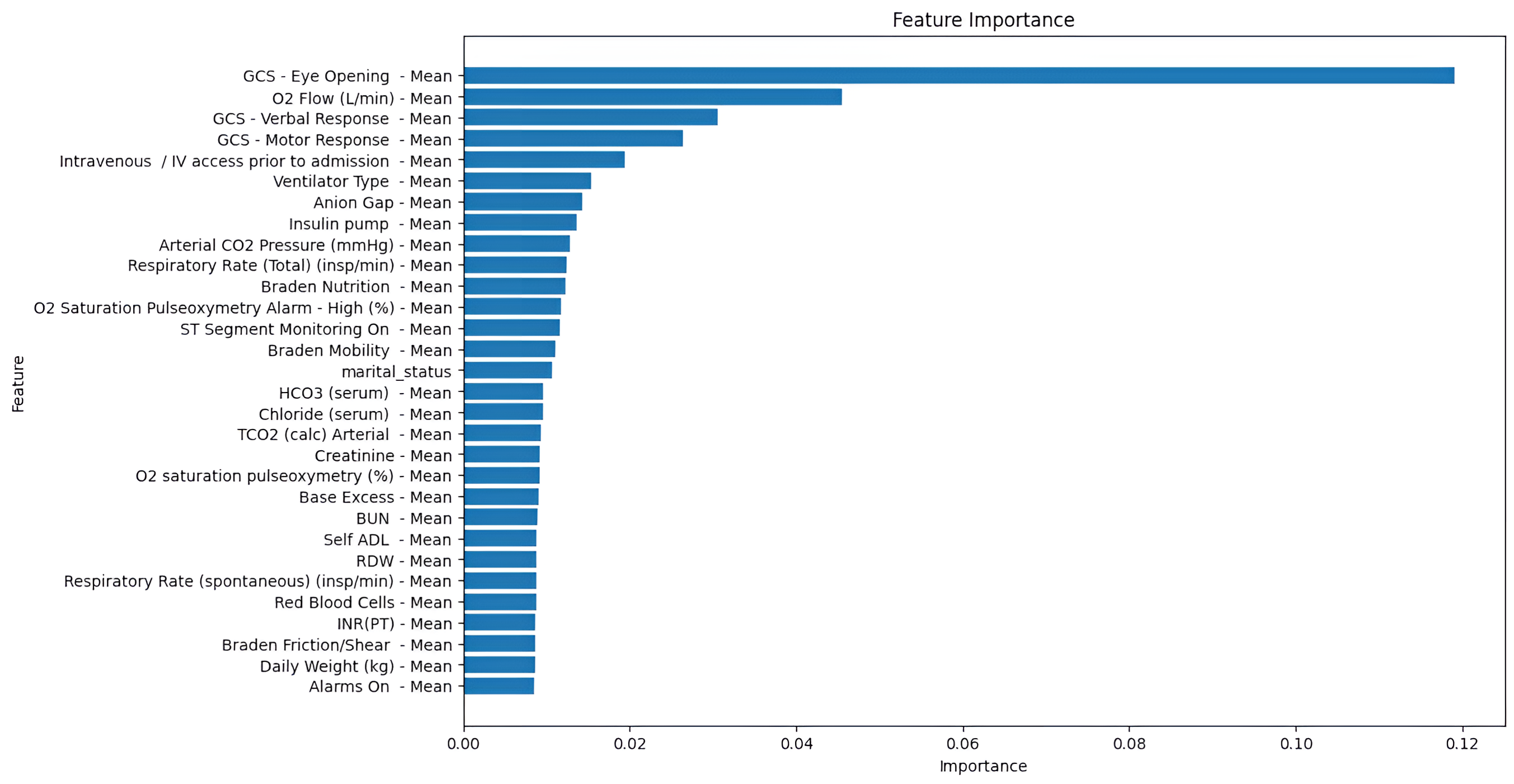} 
    \captionsetup{justification=centering}
    \caption{Feature importance and ranking based on XGBoost feature extractor.}
    \label{fig:importance}
\end{figure*}

\figref{fig:importance} shows the importance of each feature as results of XGBoost algorithm. \mytabref{tab:metrics} summarizes the performance criteria of various machine learning models designed to predict patient mortality, highlighting how each model excels or lags in specific criteria. The LASSO-RF model demonstrates exemplary sensitivity, making it highly effective at identifying patients at high risk of mortality. In contrast, the XGB-LR model boasts the highest precision, indicating its accuracy in confirming cases when a positive result is predicted. Meanwhile, the XGB-RF model balances both precision and sensitivity effectively, achieving the highest F1-score among all models. Notably, the XGB-DL model scores highest in specificity, which is crucial for reducing false positive rates. Each model presents a trade-off between these metrics, reflecting their suitability for different clinical scenarios depending on the desired outcome—whether it's avoiding false negatives or false positives.

\begin{table*}[t]
\renewcommand{\arraystretch}{1.3}
\begin{center}
\caption{Accuracy, Precision, Sensitivity, F1-Score, and Specificity values for different classifiers using XGBoost and LASSO as feature extractor}
\label{tab:metrics}
\begin{tabular}{|l|c|c|c|c|c|}
\hline
\textbf{} & \textbf{Accuracy} & \textbf{Precision} & \textbf{Sensitivity} & \textbf{F1-Score} & \textbf{Specificity} \\
\hline
XGB-RF & 0.866 & 0.865 & 0.989 & \textbf{0.923}* & 0.334 \\
\hline
XGB-LR & 0.783 & \textbf{0.954}* & 0.769 & 0.851 & 0.841 \\
\hline
XGB-XGB & 0.856 & 0.870 & 0.967 & 0.916 & 0.373 \\
\hline
XGB-LightGBM & 0.844 & 0.858 & 0.967 & 0.909 & 0.308 \\
\hline
\textbf{XGB-DL} & 0.853 & 0.939 & 0.836 & 0.884 & \textbf{0.864}* \\
\hline
LASSO-RF & 0.876 & 0.872 & \textbf{0.993}* & 0.928 & 0.370 \\
\hline
LASSO-LR & 0.796 & 0.942 & 0.799 & 0.865 & 0.787 \\
\hline
LASSO-XGB & \textbf{0.868}* & 0.874 & 0.977 & \textbf{0.923}* & 0.391 \\
\hline
LASSO-LightGBM & 0.856 & 0.863 & 0.978 & 0.917 & 0.326 \\
\hline
LASSO-DL & 0.845 & 0.918 & 0.889 & 0.903 & 0.655 \\
\hline
\end{tabular}
\end{center}
\end{table*}

Choosing the XGB-DL model for improving the prediction of patient mortality can be particularly advantageous due to its highest specificity among the evaluated models. Specificity measures the model’s ability to correctly identify true negatives, which, in this context, translates to accurately predicting patients who will not die. This is critical in clinical settings as high specificity minimizes false positives—cases where the model incorrectly predicts death.

However, the model has also produced a relatively low number of false positives (278), which is crucial for improving specificity. This low number of false positives means the model is not overly predicting deaths, helping to prevent unnecessary treatments or interventions for patients inaccurately flagged as high-risk. Moreover, the model has fewer false negatives (833) compared to true positives, indicating a robust balance in sensitivity as well.

\mytabref{tab:4} provides AUROC scores and 95\% Confidence Intervals (CI) for a predictive model that assesses patient outcomes every 8 hours across training, validation, and test datasets. In the training set, the model shows exceptional performance with an AUROC of 0.945 and a very tight confidence interval between 0.944 and 0.947, demonstrating consistent accuracy within this dataset. However, a noticeable decline in performance is observed when the model is applied to the validation and test sets, with AUROCs of 0.876 and 0.878, respectively. The slightly broader confidence intervals of 0.865-0.889 for validation and 0.866-0.888 for test indicate more variability in the model's performance on new, unseen data. This drop suggests that while the model is highly effective with training data, its generalizability is somewhat limited, possibly due to overfitting. This observation underscores the necessity for additional model tuning or adjustments in model complexity to enhance its applicability across diverse datasets.

\begin{table}[htp]
\centering
\caption{AUROC and 95\% Confidence Interval for Train, Validation, and Test Sets}
\renewcommand{\arraystretch}{1.2} 
\label{tab:4}
\begin{tabular}{|c|c|c|}
\hline
\textbf{Set} & \textbf{AUROC} & \textbf{95\% CI} \\
\hline
Train set & 0.945 & [0.944 - 0.947] \\
\hline
Validation set & 0.876 & [0.865 - 0.889] \\
\hline
Test set & 0.878 & [0.866 - 0.888] \\
\hline
\end{tabular}
\end{table}

In parallel, \mytabref{tab:5} details the performance of the same predictive model over the initial four days, highlighting a progressive improvement in its ability to accurately forecast patient outcomes. Starting with an AUROC of 0.865 on the first day, the score steadily increases to 0.903 by the fourth day. The accompanying 95\% CIs for each day's AUROC also tighten significantly by the fourth day, ranging from 0.868 to 0.936, which boosts confidence in the model's predictions as more data is analyzed over time. In the last column of this table, our results are compared with the previous study which shows a huge improvement (11-15\%) in AUROC. In the previous best model, the AUROC does not improve over time necessarily, leading to a weak model for predicting mortality over time. In contrast, our model demonstrates a significant progressive improvement, making it a valuable tool for mortality prediction over time, which is of utmost importance. Note that previous study has not reported CI for their AUROC values.

\begin{table*}[htp]
\centering
\caption{AUROC Comparison XGB-DL with previous best study with additional column showing our model 95\% Confidence Interval}
\renewcommand{\arraystretch}{1.2} 
\label{tab:5}
\begin{tabular}{|c|c|c|c|}
\hline
\textbf{Day} & \textbf{Proposed Model AUROC} & \textbf{95\% CI} & \textbf{Best Model AUROC} \\
\hline
Day 1 & 0.865 & [0.821 - 0.905] & 0.742 \\
\hline
Day 2 & 0.882 & [0.844 - 0.920] & 0.776 \\
\hline
Day 3 & 0.882 & [0.841 - 0.917] & 0.754 \\
\hline
Day 4 & 0.903 & [0.868 - 0.936] & 0.750 \\
\hline
\end{tabular}
\end{table*}

\begin{figure*}[!t]
    \centering
    \begin{minipage}{.5\textwidth}
        \includegraphics[width=0.95\linewidth]{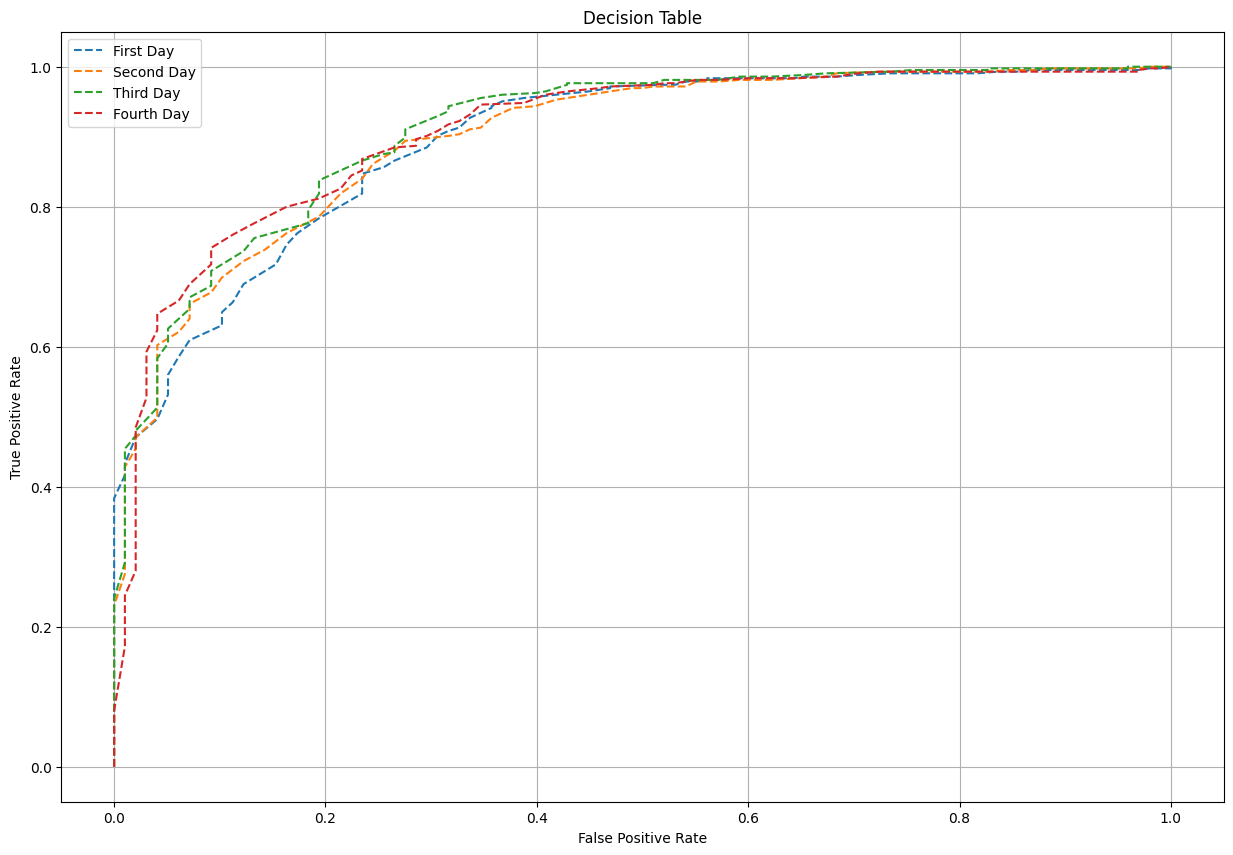}
        \centering \small \textbf{(a)} Random Forest
    \end{minipage}\hfill
    \begin{minipage}{.5\textwidth}
        \includegraphics[width=0.95\linewidth]{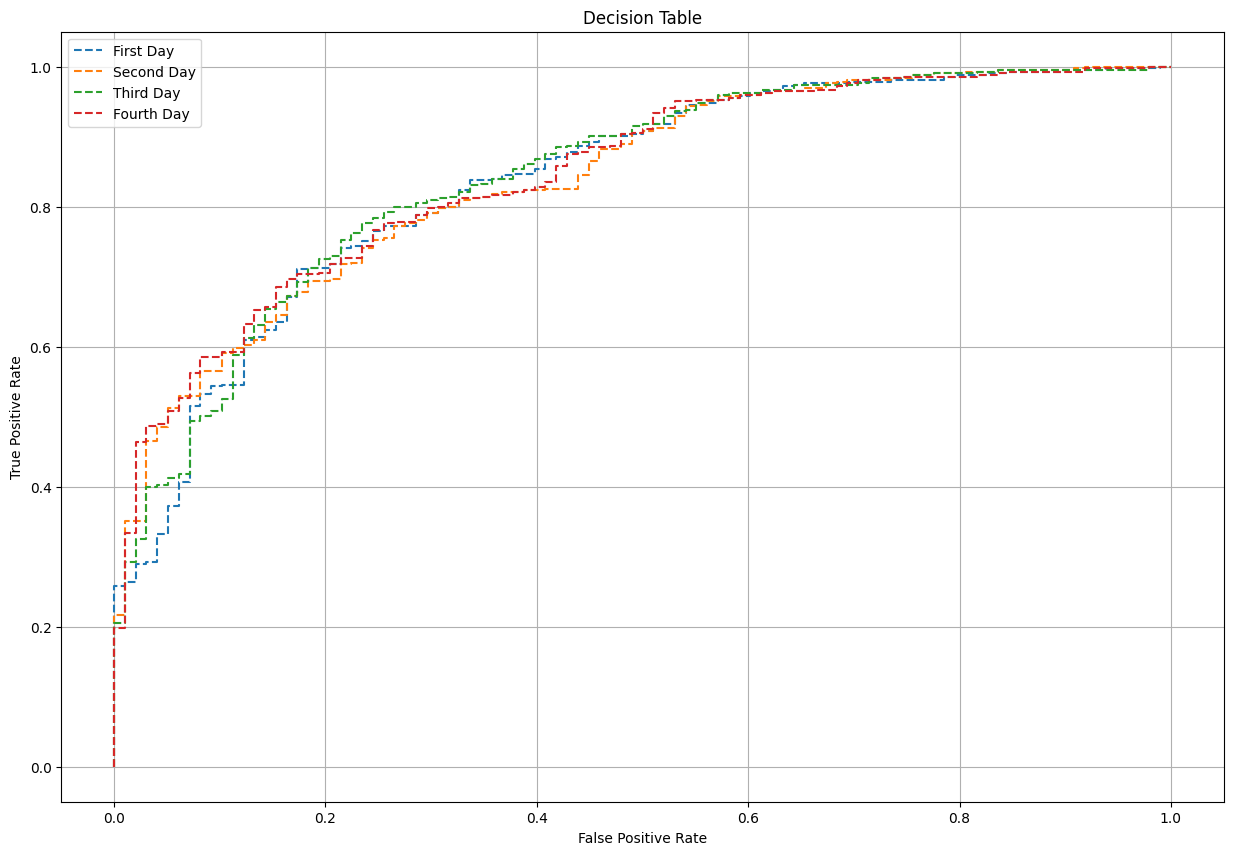}
        \centering \small \textbf{(b)} XGB
    \end{minipage}
    
    \vfill 
    
    \begin{minipage}{.5\textwidth}
        \includegraphics[width=0.95\linewidth]{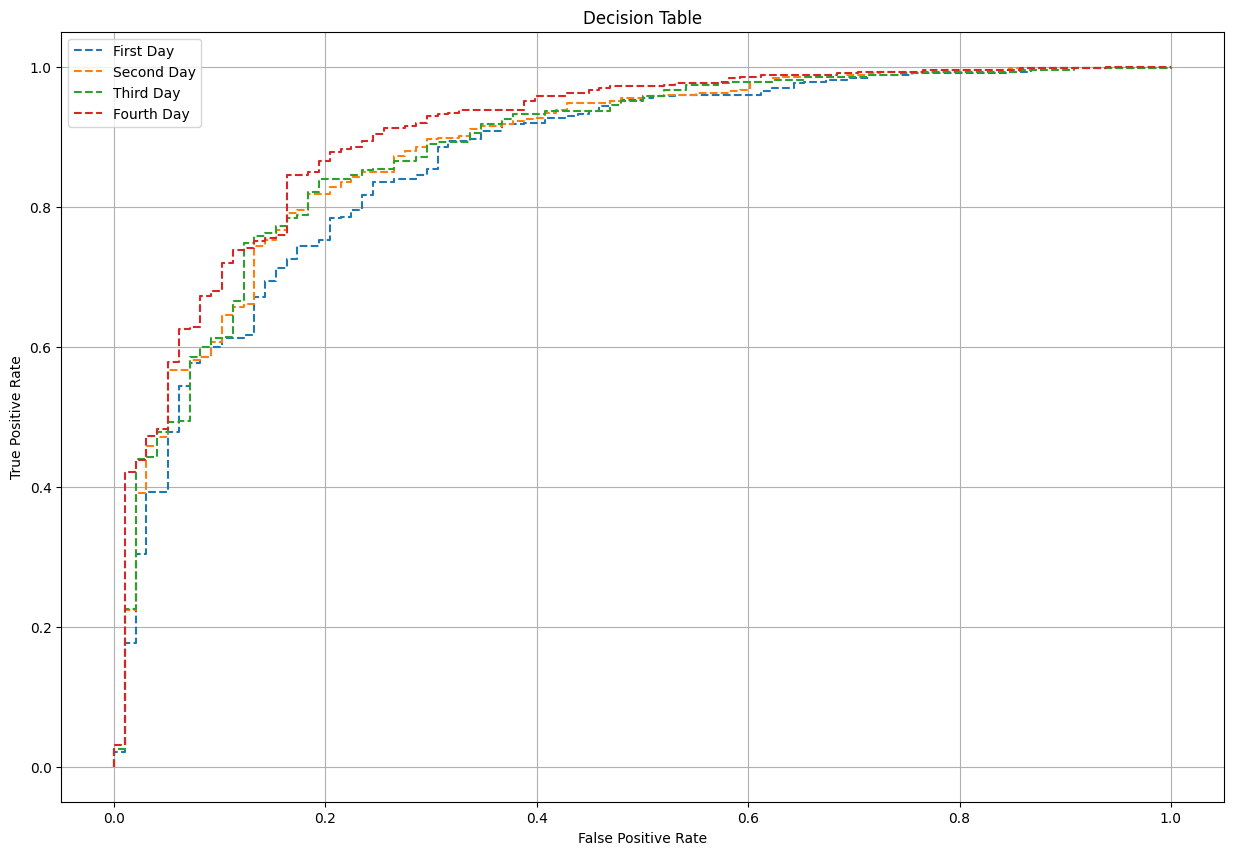}
        \centering \small \textbf{(c)} Deep Learning
    \end{minipage}\hfill
    \begin{minipage}{.5\textwidth}
        \includegraphics[width=0.95\linewidth]{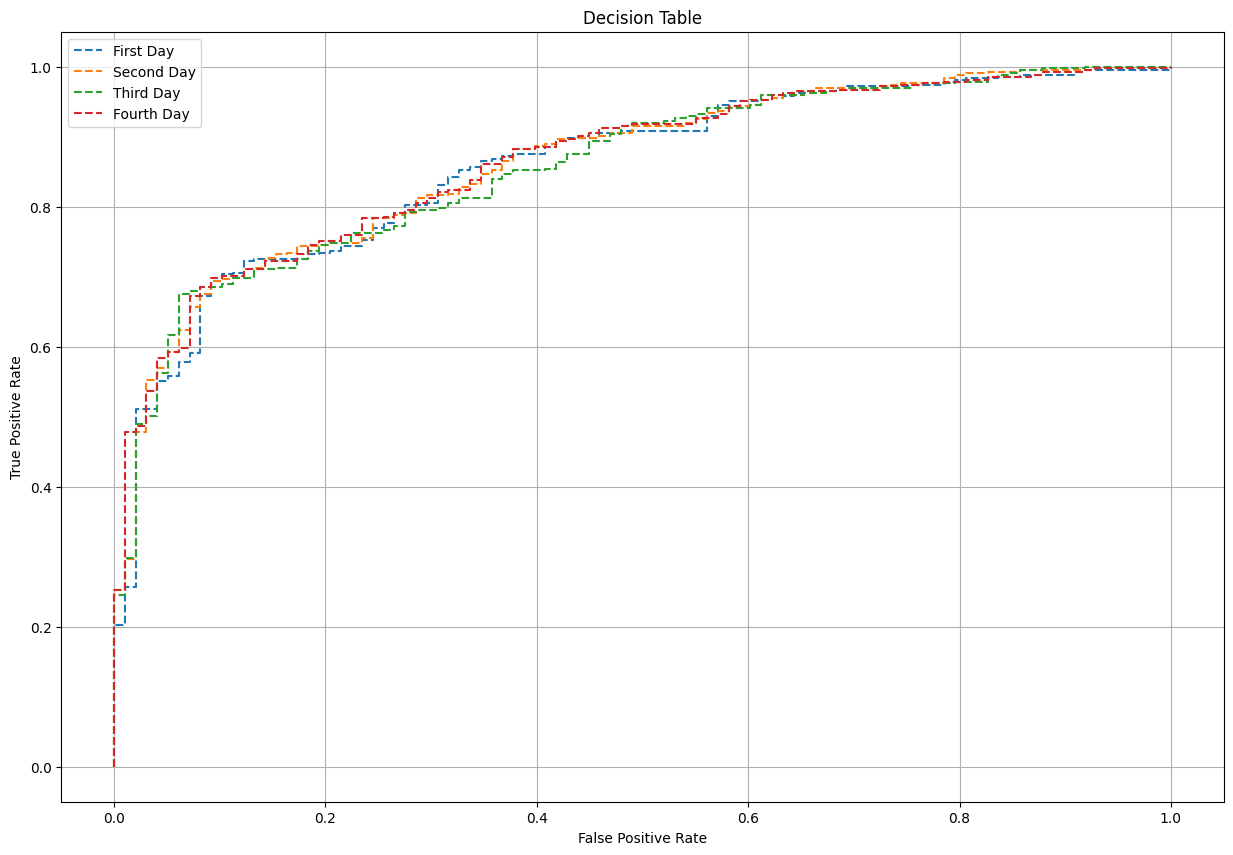}
        \centering \small \textbf{(d)} LightGBM
    \end{minipage}
    
    \vfill 
    
    \begin{minipage}{.5\textwidth}
        \includegraphics[width=0.95\linewidth]{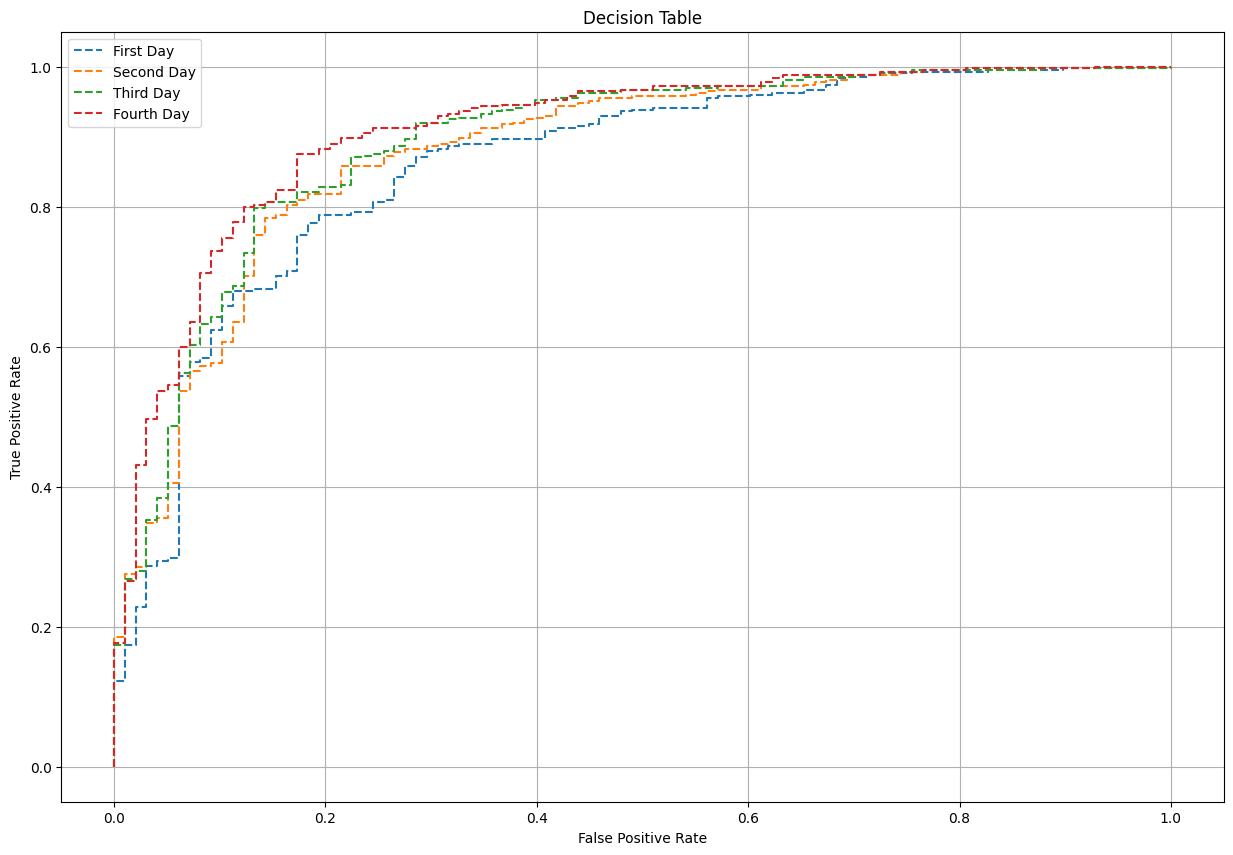}
        \centering \small \textbf{(e)} Logistic Regression
    \end{minipage}
    
    \caption{AUC Comparison of different classifiers in four days.}
    \label{fig:algorithm_comparison}
\end{figure*}

The ROC curve for different methods is plotted in \figref{fig:algorithm_comparison}.
This figure demonstrates that due to our effective feature selection, all our machine learning models outperformed the top study in terms of AUROC.
Deep learning model in this figure, climbing closer to the top-left corner, visually confirms the trend of improvement over days, indicating a significant enhancement in the model's reliability in predicting patient mortality. This trend underscores the model’s increasing effectiveness at prognosticating outcomes as it processes an expanding dataset across consecutive days. \mytabref{table:auroc_scores} Shows the AUROC and 95\% confidence interval for different implemented algorithms used in this paper. It is important to note that all of these models outperform the best study. Although Random Forest and Logistic Regression perform well in terms of AUROC, as does Deep Learning, we choose Deep Learning because of its higher specificity. 

\begin{table*}[h]
\centering
\caption{AUROC Scores with 95\% CI from Day 1 to Day 4 for different algorithms}
\renewcommand{\arraystretch}{2}
\begin{scriptsize}
\begin{tabular}{|c|c|c|c|c|c|}
\hline
 & \textbf{Random Forest} & \textbf{Logistic Regression} & \textbf{XGBoost} & \textbf{LightGBM} & \textbf{Deep Learning} \\ \hline
\textbf{Day 1} & 0.8943 [0.8603-0.9255] & 0.8604 [0.8132-0.9008] & 0.8362 [0.7898-0.8809] & 0.8590 [0.8199-0.8959] & 0.8657 [0.8214-0.9057] \\ \hline
\textbf{Day 2} & 0.8985 [0.8627-0.9263] & 0.8793 [0.8369-0.9159] & 0.8393 [0.7984-0.8780] & 0.8631 [0.8245-0.8972] & 0.8827 [0.8448-0.9208] \\ \hline
\textbf{Day 3} & 0.9092 [0.8766-0.9370] & 0.8910 [0.8509-0.9254] & 0.8427 [0.7995-0.8837] & 0.8584 [0.8251-0.8930] & 0.8825 [0.8417-0.9180] \\ \hline
\textbf{Day 4} & 0.9062 [0.8722-0.9373] & 0.9078 [0.8703-0.9388] & 0.8458 [0.8044-0.8832] & 0.8652 [0.8286-0.8968] & 0.9039 [0.8686-0.9362] \\ \hline
\end{tabular}
\end{scriptsize}
\label{table:auroc_scores}
\end{table*}

\subsection{\textit{SHAP analysis}}

\begin{figure}[!htb]
    \centering
    \includegraphics[width=1\linewidth]{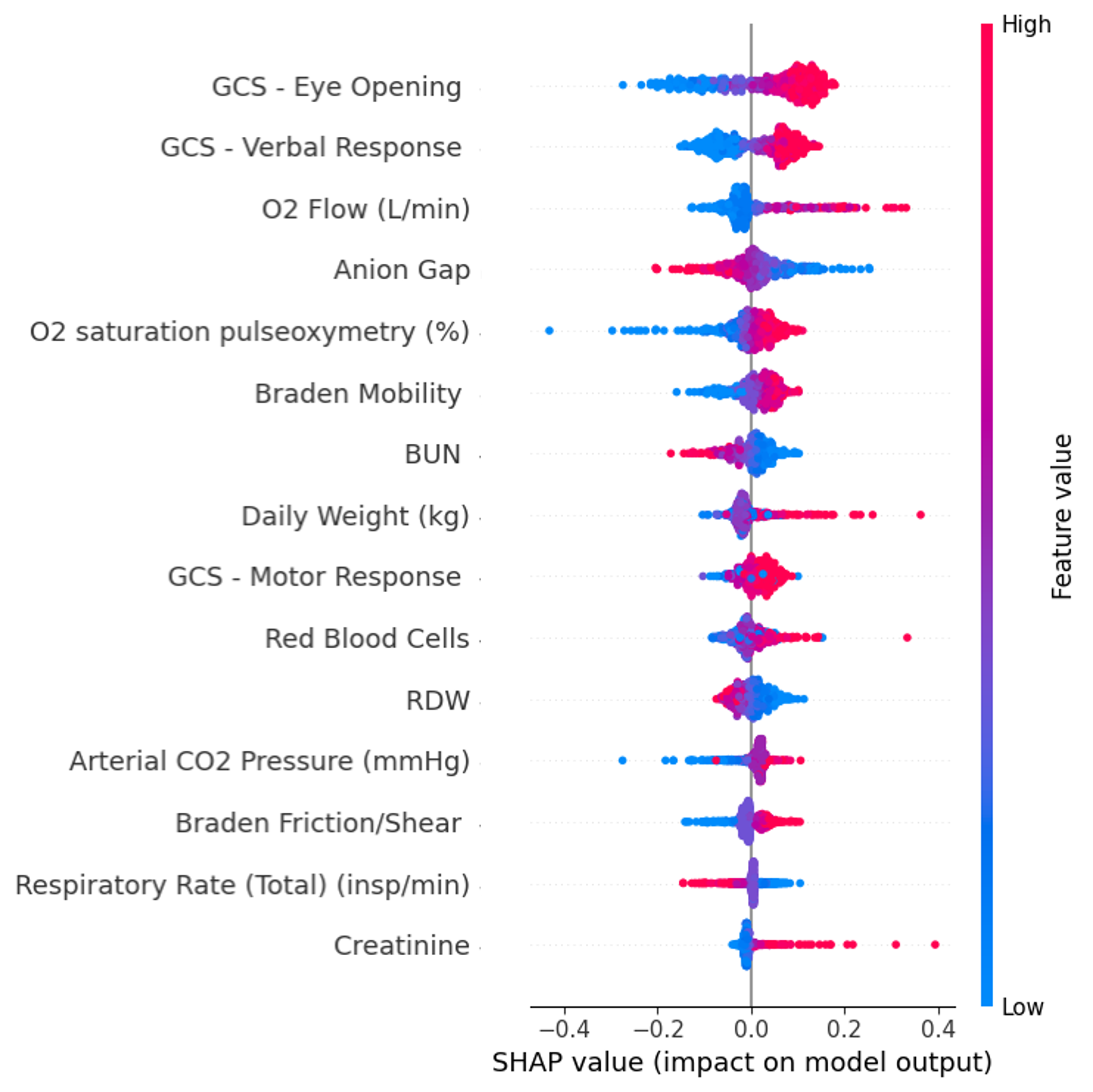}
    \caption{SHAP value based on neural network model for the test set.}
    \label{fig:6}
\end{figure}

The SHAP (SHapley Additive exPlanations) analysis graph effectively utilizes machine learning techniques to quantify and visually represent the significance of various clinical parameters in a predictive model. This analysis robustly interprets the impact of individual features on the model's predictions, enhancing our understanding of the underlying mechanisms driving outcomes \cite{ref35}. \figref{fig:6} illustrates the influence of the top 15 features on the output of the predictive model, highlighting the importance of each feature in shaping the model's predictions.

Features like the components of the Glasgow Coma Scale (GCS - Eye Opening, Verbal Response, Motor Response) predominantly show positive SHAP values, suggesting that higher scores significantly improve the model’s predictions, typically towards more favorable outcomes. Conversely, features such as O2 Flow and O2 Saturation Pulseoxymetry exhibit both positive and negative impacts, highlighting the complexity of their roles in influencing patient outcomes based on additional health contexts. Laboratory values like Anion Gap and BUN, as well as physiological measures such as Daily Weight and Creatinine, display varied impacts, indicating their roles in assessing the severity of patient conditions and organ function. Features like Respiratory Rate and Arterial CO2 Pressure add depth to the nuanced understanding of how respiratory health affects the model. This SHAP analysis is instrumental in unraveling the direct and interactive effects of various clinical parameters on model behavior, fostering more precise enhancements to model accuracy and clinical relevance. By providing clear insights into how individual features affect mortality predictions, SHAP enables healthcare providers to make more informed and targeted decisions. For example, understanding that higher GCS scores are associated with lower mortality risk allows clinicians to quickly assess neurological function and tailor interventions accordingly. These insights not only improve the transparency of the model but also build trust among clinicians, as they can clearly see how specific patient characteristics influence predictions.

\section{DISCUSSION}

\subsection{\textit{Existing model compilation summary}}

This study successfully developed a deep learning approach that significantly enhances the prediction of mortality among ICU patients suffering from ischemic stroke. Compared to best-existing literature \cite{ref17}, all of our models improved the AUROC significantly and the baseline model demonstrates 13\% improvement on average on AUROC by utilizing a carefully curated set of 30 features, a substantial reduction from \cite{ref17} which uses 1095 features. we achieved higher accuracy using a model on a dataset with more than 30 times fewer features. This remarkable result underscores the effectiveness of our innovative feature selection techniques and the robustness of our modeling approach. By drastically reducing the feature set, we not only simplified the model but also enhanced its performance and generalization capabilities. Also, this leads to a lot of calculation reduction which makes this model much faster.

One of the standout features of the XGB-DL model is its specificity, which reaches up to 86.4\% in distinguishing true negatives. This aspect is crucial in the clinical environment, where accurate prediction of patient outcomes can significantly influence treatment decisions and resource allocation. Moreover, the model’s AUROC improved progressively from 86.5\% (CI 82.1\% - 90.5\%) to 90.3\% (CI 86.8\% - 93.6\%) over the first four days of patient admission, indicating increasing predictive accuracy that could be pivotal for clinical interventions during critical early stages.

\subsection{\textit{Study limitations}}

However, the practical deployment of such a sophisticated model in diverse ICU settings may encounter challenges, including the need for integration into existing medical record systems and the potential requirement for staff training on new technologies. 

One limitation of this study is that it does not account for patients with recurrent ICU admissions due to chronic conditions or complications. Future research could explore the impact of including multiple ICU admissions to assess how recurrent ICU stays affect mortality prediction.

This study also faces other limitations due to its exclusive reliance on the MIMIC-IV database, potentially affecting the model’s applicability across different global healthcare settings. The dataset was limited, and to evaluate whether our model generalizes effectively, it is critical to incorporate additional datasets from diverse sources. Future works should focus on expanding the dataset to include records from various geographical locations and healthcare systems to ensure broader applicability and robustness. Additionally, the current imputation methods used for handling missing data can be improved. Enhancing these imputation techniques will result in more accurate and useful datasets, ultimately leading to better model performance and reliability. Future research should explore advanced imputation methods to maximize data utility and model 

\section{CONCLUSION}

This research significantly advances predictive modeling of mortality in ischemic stroke patients within ICU settings. Although Random Forest and Logistic Regression also performed well in terms of AUROC, we chose Deep Learning because of its higher specificity. The XGB-DL model, with its high specificity and improved predictive accuracy over time, promises to be a valuable tool for clinicians, enhancing patient outcomes and optimizing ICU resource utilization. Notably, our approach achieved an impressive 13\% increase in AUROC on average, while utilizing 30 times fewer features, demonstrating the model's efficiency and effectiveness.

Future studies should aim to validate this model across varied healthcare databases to ascertain its effectiveness and adaptability across different patient demographics and treatment protocols. Additionally, exploring the integration of this predictive model into clinical practice could provide insights into operational challenges and benefits, paving the way for broader adoption and potentially transforming ICU patient care management.

\section*{Acknowledgment}

The authors extend their gratitude to the creators of MIMIC-IV for furnishing a thorough and inclusive public electronic health record (EHR) dataset.

\bibliographystyle{unsrt}
\bibliography{ref}

\end{document}